\documentclass[conference]{IEEETran}
\usepackage{authblk}
\usepackage[utf8]{inputenc} 
\usepackage[T1]{fontenc}    
\usepackage{hyperref}       
\usepackage{url}            
\usepackage{booktabs}       
\usepackage{amsfonts}       
\usepackage{nicefrac}       
\usepackage{microtype}      
\usepackage{xcolor}         
\usepackage{amsmath,amssymb}
\usepackage{wrapfig}
\usepackage{stfloats}
\usepackage{graphicx}
\usepackage{float}
\usepackage{caption}
\usepackage{subcaption}
\usepackage{bm}
\usepackage{listings}
\usepackage{multicol}

\NewDocumentCommand{\codeword}{v}{%
\texttt{\textcolor{blue}{#1}}%
}

\title{A Temporal Kolmogorov-Arnold Transformer for Time Series Forecasting}

\author[1]{Rémi Genet}
\author[2]{Hugo Inzirillo}

\affil[1]{DRM, Université Paris Dauphine - PSL}
\affil[2]{CREST-ENSAE, Institut Polytechnique de Paris}

\begin{document}

\maketitle
\thispagestyle{plain}
\pagestyle{plain}
\begin{abstract}
Capturing complex temporal patterns and relationships within multivariate data streams is a difficult task. We propose the Temporal Kolmogorov-Arnold Transformer (TKAT), a novel attention-based architecture designed to address this task using Temporal Kolmogorov-Arnold Networks (TKANs). Inspired by the Temporal Fusion Transformer (TFT), TKAT emerges as a powerful encoder-decoder model tailored to handle tasks in which the observed part of the features is more important than the a priori known part. This new architecture combined the theoretical foundation of the Kolmogorov-Arnold representation with the power of transformers. TKAT aims to simplify the complex dependencies inherent in time series, making them more "interpretable". The use of transformer architecture in this framework allows us to capture long-range dependencies through self-attention mechanisms.

\begin{figure}[H]
        \centering
    \includegraphics[width=1\linewidth]{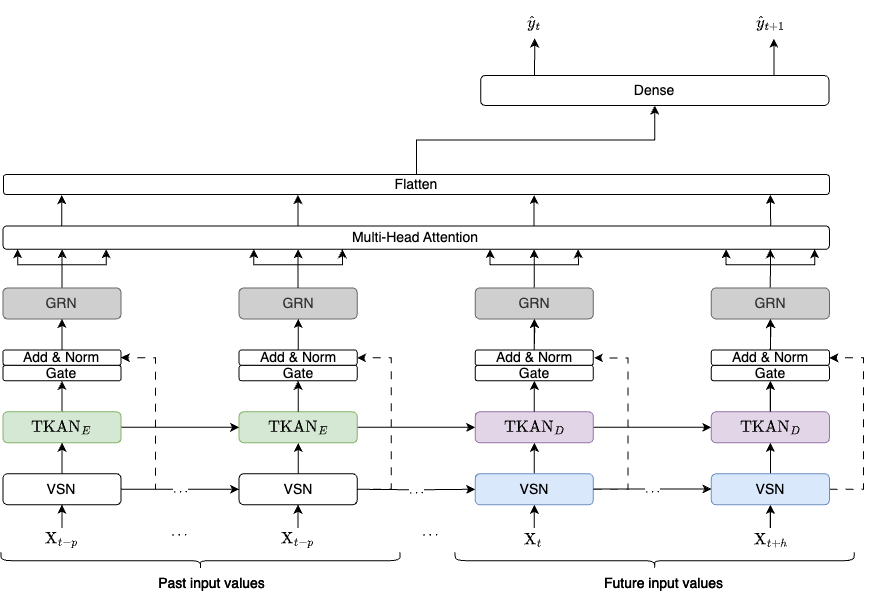}
    \caption{Temporal Kolmogorov-Arnold Transformer (TKAT)} 
    \label{fig:temporal_kan}
\end{figure}

\end{abstract}
\section{Introduction}
Multivariate, time series forecasting is an area of strong interest for researchers \cite{taieb2010multiple,tiao1989model,wei2018multivariate,marcellino2006comparison} and has become an important area of research for many industries. The volume of data available is increasing, in some way it is natural to propose more advanced frameworks for prediction. In contrast to univariate time series, in which a single variable is analyzed over time, multivariate time series involve several interdependent variables. These connections makes the learning task even more complex. Multivariate time series data require sophisticated models that can capture dynamic interactions and temporal dependencies across multiple dimensions and time horizon. Researchers have explored various methods, including vector autoregressive models \cite{zivot2006vector,lutkepohl2013vector}, state-space models \cite{hamilton1994state,kim1994dynamic}, and neural networks \cite{tang1991time}, to tackle the challenges associated with multivariate time series forecasting to detect connections between time series \cite{binkowski2018autoregressive,ma2019novel}. Deep learning approaches such as Long Short-Term Memory (LSTM) networks and attention mechanisms have shown promising results due to their ability to handle non-linearities and long-range dependencies in data \cite{hochreiter1997long, vaswani2017attention}. These advanced techniques are crucial for improving forecasting accuracy and making informed decisions based on the predicted outcomes.

\begin{figure}[H]
    \centering
    \includegraphics[width=1\linewidth]{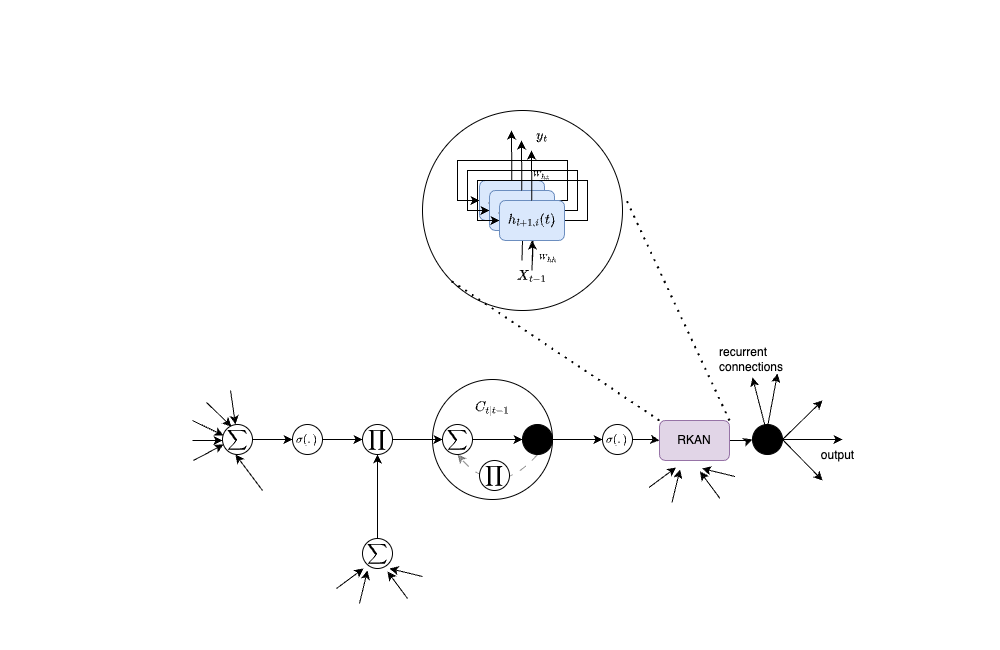}
    \caption{Temporal Kolmogorov-Arnold Networks (TKAN)} 
    \label{fig:temporal_kan}
\end{figure}

To improve forecasting accuracy, some researchers have suggested employing attention mechanisms \cite{shih2019temporal} or convolutional methods for multivariate time series prediction\cite{wan2019multivariate}. More recently,  Liu et al. \cite{liu2024kan} released the Kolmogorov-Arnold Netowrks (KANs) a promising architecture, introduced as an alternative for MLPs. To dive into the architecture of the model, we redirect the reader to \cite{liu2024kan}. KANs outperforms conventional MLPs in real-world forecasting task \cite{vaca2024kolmogorov}. Even more recently, we introduced an extension, incorporating time dependency as well as memory management \cite{genet2024tkan}. Other extension using wavelets \cite{bozorgasl2024wav} have been proposed in the mean time. TKANs aimed to develop a framework with two key functionalities: handling sequential data and managing memory. To do so, \cite{genet2024tkan} enrich the initial model adding a recurring layer of KANs to introduce RKANs. Temporal Kolmogorov-Arnold Networks (TKANs) is an upgrade version of the LSTM \cite{hochreiter1997long} using Kolmogorov-Arnold Networks (KANs). It uses RKANs for short term memory as well as a cell state, for further details we refer the read to \cite{genet2024tkan}. While TKAN layers stacked in TKANs are capable to detect temporal dependencies on time series data, it seems natural to propose an alternative to the TFT \cite{lim2021temporal} using TKAN layers instead of LSTM layers. Codes are available at \href{https://github.com/remigenet/TKAT}{https://github.com/remigenet/TKAT} and can be installed using the following command: \codeword{pip install tkat}. 
The data are accessible if you wish to reproduce the experiments inside the github provided above.
\section{Related Work}
Time series forecasting is a highly complex task. Managing depency within a time series have been the biggest challenge for many industries. RNNs appeared to be the most approtate model to tackle dedepency problem. In 2017, Transformer \cite{vaswani2017attention} have been introduced to tackle issues in
natural language processing (NLP), translation \cite{mehta2020delight,raffel2020exploring,so2019evolved,fan2021mask}, language modeling \cite{shoeybi2019megatron,dai1901attentive,roy2021efficient,rae2019compressive} and named entity recognition\cite{yan2019tener,arkhipov2019tuning,yu2020improving}.
A Transformer-based approach have been introduced \cite{lim2021temporal,wu2020deep} to forecast time series data. Compared to other sequence-aligned deep learning methods, this architecture is leveraging self-attention mechanisms to model sequence data and make possible to learn complex dependencies on various lengths of time series data. Transformer architecture has increase the qualitiy of prediciton there are several severe issues with this architecture that prevent it from being directly applicable to Long-Term Sequence Forecasting (LSTF), including quadratic time complexity, high memory usage, and inherent limitation of the encoder-decoder architecture. To overcome this problem \cite{zhou2021informer} introduced an efficient transformer-based model for LSTF. To address the issue of memory usage, \cite{li2019enhancing} porposed a LogSparse Transformer, which has a memory cost of just $O(L(log L)^{2})$. This model enhances forecasting accuracy for time series data with fine granularity and significant long-term dependencies, all while operating under a constrained memory budget. Another transformer based model is the Adversarial Sparse Transformer(AST)\cite{wu2020adversarial}, a novel architecture for time series forecasting. They propose Sparse Transformer to improve the ability to pay more attention on relevant steps in time series. By adversarial learning, they improve the contiguous and fidelity at the sequence level.To address these challenges, we propose the Temporal Kolmogorov-Arnold Transformer (TKAT), a novel architecture inspired by the temporal fusion transformer \cite{lim2021temporal} which integrates the theoretical foundation of the Kolmogorov-Arnold representation \cite{kolmogorov1961representation} with the power of transformers. TKAT aims to simplify the complex dependencies inherent in time series, making them more "interpretable". The use of transformers architecture in this framework allows us to  capture long-range dependencies through self-attention mechanisms.

\medskip

\textit{In this paper, we will propose a novel approach for temporal transformers. We will use the Multi-Head Attention mechanism \cite{vaswani2017attention}, but with  updates on encoder-decoder phases and temporal processing. Our approach aims to integrate the effective architecture of the TFT with the TKAN, which has demonstrated improved performance in n-step ahead forecasting.}

\section{Model Architecture}
We propose a new transformer architecture using the layers of the Temporal Kolmogorov-Arnold Network \cite{genet2024tkan} inspired by the Temporal Fusion Transformer (TFT)\cite{lim2021temporal}, the general idea is to use TKAN instead of LSTM for the encoding and decoding layers, but also to propose an architecture adapted to problems where known inputs are not in the majority while observed inputs are, which is typically the case for financial tasks. This is a major difference from the standard task for which the TFT was designed, and we have observed here that for series with few known inputs, our architecture offers much better performance.

\subsection{TKAN}
\label{sec:tkan}

Recurrent Kernel is the key for RKAN layers learn from sequence where context and order of data matters. We have designed the TKAN to leverage the power of Kolmogorov-Anrold Network while offering memory management to handle time dependency. To introduce time dependency each transformation function \(\phi_{l,j,i}\) to be time dependent. Let us denote \(h_{l,i}(t)\) have been modify to incorporate a "memory function" capturing the history of node  $i$ in l-$th$ layer:

\begin{equation}
x_{l+1,j}(t) = \sum_{i=1}^{n_l} \tilde{x}_{l,j,i}(t) = \sum_{i=1}^{n_l} \phi_{l,j,i,t}(x_{l,i}(t), h_{l,i}(t)),
\end{equation}
Where $j=1,\cdots,n_{l+1}$. The "memory" step \(h_{l,i}(t)\) is defined as a combination of past hidden states, such:

\begin{equation}
h_{l,i}(t) =  W_{hh} h_{l,i}(t-1) + W_{hz} x_{l,i}(t),
\label{eq:update_state_tkan}
\end{equation}

About the Recurring KAN Layer, the network now incorporate memory management at each layer:
\begin{equation}
\mathrm{KAN}(x, t) = (\mathbf{\Phi}_{L-1,t} \circ \mathbf{\Phi}_{L-2,t} \circ \cdots \circ \mathbf{\Phi}_{1,t} \circ \mathbf{\Phi}_{0,t}) (x, t).
\label{eq:rkan}
\end{equation}
The major components of the architecture of TKANs:
\begin{itemize}
    \item \textbf{RKAN Layers:} RKAN layers enable the retention of short term memory from previous states within the network. Each RKAN layer manages this short term memory throughout the processing in each layer.
    \item \textbf{Gating Mecanism:} these mechanisms help to manage the information flow. The model decides which information should be retained or forgotten over time.
\end{itemize}
For the next step, towards to maintain the memory, we took inspiration from the LSTM \cite{hochreiter1997long,staudemeyer2019understanding}.  We denote the input vector of dimension $d$ denoted by $x_t$. This unit uses several internal vectors and gates to manage information flow. The forget gate, with activation vector $f_t$, 
\begin{equation}
    f_t = \sigma(W_f x_t + U_f h_{t-1} + b_f),
\end{equation}
decides what information to forget from the previous state. The input gate, with activation vector denoted $i_t$, 
\begin{equation}
    i_t = \sigma(W_i x_t + U_i h_{t-1} + b_i),
\end{equation}

controls which new information to include. The output gate, with activation vector $o_t$,

\begin{equation}
    o_t = \sigma(\text{KAN}(\Vec{x}, t)),
\end{equation}
determines what information from the current state to output given $\text{KAN}(\Vec{x}, t)$ from Eq. \eqref{eq:rkan} . The hidden state, $h_t$, captures the unit's output, while the cell state, $c_t$ is updated such

\begin{equation}
    c_t = f_t \odot c_{t-1} + i_t \odot \tilde{c}_t,
\end{equation}
where $\tilde{c}_t = \sigma(W_c x_t + U_c h_{t-1} + b_c)$
represents its internal memory. All these internal states have a dimensionality of $h$.
The ouput denoted $h_t$ is given by

\begin{equation}
    h_t = o_t \odot \tanh(c_t).
    \label{eq:hidden_update}
\end{equation}

\subsection{Encoder-Decoder}
Encoding and Decoding task allows sequence-to-sequence mapping (ie: machine translatation, text mining, etc.). Picking the usecase of machine translation, the global idea of the task is to convert an input sequence into target sequence. In the TKAT, we use layers of TKAN detailed section \ref{sec:tkan} for both encoding and decoding tasks.

\subsubsection{Encoder}
The encoder is composed of single or multiple layers of $\text{TKAN}_E$. The recurring mecanism within the layer allows to monitor the information and to keep specific part of the sequences, and so, to capture long range dependencies. The encoder will be fed by the past inputs (observed/known inputs) filtered by the variable selection networks (VSN) detailed in a later section.

\subsubsection{Decoder}

Like the encoder, the decoder consists of several layers of $\text{TKAN}_D$. Each layer allow to process the information while maintening the memory and time dependency. However, the decoder will take as an initial state the encoder final cell state.

\subsection{Gated Residual Networks}

The relationship between temporal data is a key issue. Gated residual networks offer an efficient and flexible way of modelling complex relationships in data. They allow to control the flow of information and facilitate the learning tasks. They are particularly useful in areas where non-linear interactions and long-term dependencies are crucial. In our model we use the Gated Residual Networks proposed by \cite{lim2021temporal}, we kept the same architecture. However, we remove the need of the context.

\begin{figure}[H]
    \centering
    \includegraphics[width=0.7\linewidth]{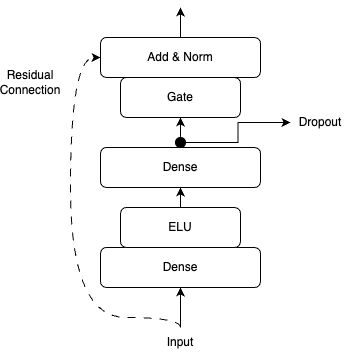}
    \caption{Gated Residual Networks (GRN) \cite{lim2021temporal}} 
    \label{fig:vsn}
\end{figure}

\begin{align}
\text{GRN}_\omega\left(x \right) &=\text{LayerNorm}\left(x + \text{GLU}_\omega(\eta_1) \right), \\
    \eta_1 &= W_{1, \omega}~\eta_2 + b_{1, _\omega},  \label{eqn:grn_step}\\
    \eta_2 &= \text{ELU}\left( W_{2, \omega}~x + b_{2, _\omega} \right), & \label{eqn:grn_base}
\end{align}
In this context, $\text{ELU}$ designates the Exponential Linear Unit activation function \cite{clevert2015fast}, while $\eta_1 \in \mathbb{R}^{d_{model}}$ and $\eta_2 \in \mathbb{R}^{d_{model}}$ represent intermediate layers. The standard layer normalization $\text{LayerNorm}$ is that described in \cite{ba2016layer}, and $\omega$ is an index used to indicate weight sharing. When the expression $W_{2, \omega} x + b_{2, \omega}$ is largely positive, ELU activation works as an identity function. On the other hand, when this expression is largely negative, ELU activation produces a constant output, thus behaving like a linear layer. We use Gated Linear Units (GLUs) \cite{dauphin2017language} to provide the flexibility to suppress any parts of the architecture that are not required for a given dataset. Letting $\gamma \in \mathbb{R}^{d_{model}}$ be the input, the GLU then takes the form:
\begin{align}
 \text{GLU}_\omega(\gamma) & =  \sigma(W_{4, \omega}~\gamma + b_{4, \omega}) \odot (W_{5, \omega}~\gamma + b_{5, \omega} ),
\label{eqn:component_gate}
\end{align}
where $\sigma(.)$ is the sigmoid activation function, 

$W_{(.)} \in \mathbb{R}^{d_{model}\times d_{model}}, b_{(.)} \in \mathbb{R}^{d_{model}}$ are the weights and biases, $\odot$ is the element-wise Hadamard product, and $d_{model}$ is the hidden state size (common across TFT).
GLU allows TFT to control the extent to which the GRN contributes to the original input $x$ -- potentially skipping over the layer entirely if necessary as the GLU outputs could be all close to 0 in order to suppress the nonlinear contribution.

\subsection{Variable Selection Networks}

\begin{figure}[H]
    \centering
    \includegraphics[width=0.8\linewidth]{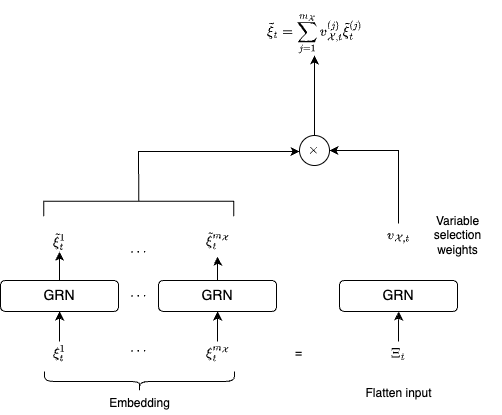}
    \caption{Variable Selection Networks (VSN)} 
    \label{fig:vsn}
\end{figure}

The Variable Selection Networks (VSN), will help model performance via utilization of learning capacity only on the most salient covariates. Figure \ref{fig:vsn} described the different step in the VSN. We define ${\xi}_{t}^{(j)}  \in \mathbb{R}^{d_{model}}$  the transformed input of the $j$-th variable at time $t$, with ${\Xi}_t = \left[ {\xi}_{t}^{(1)^T}, \dots, {\xi}_{t}^{(m_\chi)^T}  \right]^T$  the flattened vector of all past inputs at time $t$. Variable selection weights denoted $v_{\mathcal{X},t}$  generated by feeding both ${\Xi}_t$ and an external context vector ${c}_{s}$ through a GRN and followed by a Softmax layer such: 
\begin{align}
{v}_{\chi_t} = \text{Softmax}\left(\text{GRN}_{v_{\chi}}({\Xi}_t)\right),& 
\label{eq:vsn}
\end{align}
where ${v}_{\chi_t} \in \mathbb{R}^{m_\chi}$ is a vector of variable selection weights. We remove the need of a static covariate encoder as we are not embedding static covariate within our model. For each time step, a non-linear processing is employed by feeding each ${\xi}_{t}^{(j)}$ through its own GRN:
\begin{align}
\tilde{{\xi}}_t^{(j)} = \text{GRN}_{\tilde{\xi}(j)}\left({\xi}_t^{(j)}\right), & \label{sec:varselect_grn}
\end{align}
where $\tilde{{\xi}}_t^{(j)}$ is the processed feature vector for variable $j$. We note that each variable has its own $\text{GRN}_{\xi(j)}$, with \emph{weights shared across all time steps $t$}. Processed features are then weighted by their variable selection weights and combined:
\begin{align}
\tilde{{\xi}}_t  = \sum\nolimits_{j=1}^{m_\chi} v_{\chi_t}^{(j)}  \tilde{{\xi}}_t^{(j)}, &
\label{eq:var_selection_sum}
\end{align}
where $v_{\chi_t}^{(j)}$ is the j-th element of vector  ${v}_{\chi_t} $.

\subsection{Temporal Decoder}
The temporal decoder in the Temporal Fusion Transformer generates accurate and interpretable time series forecasts. This is mainly due to using attention mechanisms, positional encodings, static context, and recurrent neural units to capture short-term and long-term dependencies. The TKAT will also embeds a temporal decoder to make accurate forecast, however the structure of the temporal decoder will be slightly different.

\subsubsection{Self-Attention Layer}
The TKAT uses a self-attention mechanism to capture long-term relationships between different time steps modified from multi-head attention in transformer-based architectures \cite{li2019enhancing}.

\begin{align}
\text{Attention}(Q, K, V) &= A(Q, K) V, 
\label{attn}
\end{align}
where $A(.)$ is a normalization function. A common choice is the scaled dot-product attention ~\cite{vaswani2017attention}:
\begin{align}
A(Q, K) = \text{Softmax} \left( \frac{Q K^T}{\sqrt{d_{attn}}} \right).
\label{attn_normalize}
\end{align}

To enhance the learning capacity of the standard attention mechanism, \cite{vaswani2017attention} proposed multi-head attention, which employs different heads for different representation subspaces:
\begin{align}
\text{MultiHead} (Q, K, V) &= [H_1, \dots,  H_{m_H}] W_H, \\
H_h &= \text{Attention}(Q W_{Q}^{(h)},  K W_{K}^{(h)}, V W_{V}^{(h)}), 
\end{align}
where \(W_{K}^{(h)} \in \mathbb{R}^{d_{model} \times d_{attn}}\), \(W_{Q}^{(h)} \in \mathbb{R}^{d_{model} \times d_{attn}}\), and \(W_{V}^{(h)} \in \mathbb{R}^{d_{model} \times d_{V}}\) are head-specific weights for keys, queries, and values, respectively. The matrix \(W_H \in \mathbb{R}^{(m_H \cdot d_{V})\times d_{model}}\) linearly combines outputs concatenated from all heads \(H_h\). Given that different values are used in each head, attention weights alone would not indicate a particular feature's importance.

\subsubsection{Fully Aware Layer}
In time series analysis, past values play a very important role in establishing reliable predictions. Historical data can capture trends, seasonality and the underlying patterns that will impact the evolution of a variable through time. Past observations also allows us to establish relation ship and interdependences. In some cases it will be possible to integrate the presence of “shocks” and find the origin or the determinants. This led us to modify this layer by proposing to flatten the output of the Temporal self-attention layer. This modification allows the network to be fully connected to past values and not just to the unknown inputs fed by the encoder's final cell state. This architecture enables us to maintain a more persistent memory. 
Given the output of the temporal self-attention layer, $m_H$ vectors, $[\tilde{H}_1,\tilde{H}_2, \dots,  \tilde{H}_{m_H}]$, where each $\tilde{H}_i$ is a vector in $\mathbb{R}^{m_i}$ and $ i=1,\dots,H$, the flattened vector \(\tilde{H}_{\text{flat}}\) in \(\mathbb{R}^{m_1 + m_2 + \dots + m_H}\) can be written as:

\begin{equation}
   \tilde{H}_{\text{flat}} = \begin{bmatrix} \tilde{H}_1 \\ \tilde{H}_2 \\ \vdots \\ \tilde{H}_{m_H} \end{bmatrix}
\end{equation}

\subsection{Ouputs}
The ouput of the TKAT is obtained from the flatten vector of MultiHeadAttention(.):

\begin{equation}
    \hat{y}_{t:t+\tau}= W_{\hat{y}}  \tilde{H}_{\text{flat}} +b_{\hat{y}}  
\end{equation}

where $ W_{\hat{y}} \in \mathbf{R}^{\tau}$The architecture proposed provides a solid framework to forecast multivariate time series data n-step ahead. The fully connected linear layer ouput dimension correspond to the number of step $k=[1,\dots,\tau]$ to forecast. The output vector $\hat{y}_{t:t+\tau}=[\hat{y}_{t+1},\dots,\hat{y}_{t+\tau}]$ contains all the forecasting values for each timestep. Thus, each predicted value is a weighted sum of all the attention results, with different weights for each step.

\section{Learning Task}
To test the model's ability to offer better predictions with our approach, we used the same tasks as in the article \cite{genet2024tkan}, for an easy comparison for the results. 

\subsection{Task definition and dataset}

The task we use consists in predicting the notional amount traded on the market over several steps ahead. We chose this task because it is far from easy, as market data is notorious for its lack of predictability, even though volume is a process that has predictable components such as seasonality over the day and week, as well as strong auto-correlation. 

\medskip

The tasks focus solely on the Binance exchange, so our dataset only contains data from this exchange, which has been the most important market for many years. We have also only used USDT markets, as this is the most widely used stablecoin, and all notionals are therefore intended in USDT.

\medskip

Our dataset consists of the notional amounts traded each hour on several assets:  BTC, ETH, ADA, XMR, EOS, MATIC, TRX, FTM, BNB, XLM, ENJ, CHZ, BUSD, ATOM, LINK, ETC, XRP, BCH and LTC, which are to be used to predict just one of them, BTC. The data period runs over 3 year from January 1, 2020 to December 31, 2022. However, all these values are observed values, i.e. they are not known in the futures, whereas, as can be seen, the model architecture requires known inputs in the futures. Thus, for the transformer, we have added 2 sets of data containing the time of day and the day of the week, which are known values and therefore present on both past and future inputs.

\subsection{Data Preparation}
Data preparation is a necessary step for most machine learning methods, to facilitate gradient descent when data are of different sizes, to ensure series stationarity, etc. This is very important on our dataset for two reasons. This is very important on our dataset for two reasons, the first being that the assets traded are at very different scales, and the second being that the notionals of an asset are not at all stationary over time.

\medskip

In order to obtain data that can be used for training, but also return a meaningful loss to optimize, we use a two-step scaling.The first step is to divide the values in the series by the moving median of the last two weeks. This moving median window is also shifted by the number of steps forward we want to predict, so as not to include foresight. This first pre-treatment aims to make the series more stationary over time. The second pre-processing applied is a simple MinMaxScaling per asset, even if here the minimum of the series is 0, it is simply a matter of dividing by the maximum value. The aim is to scale the data in the 0, 1 interval to avoid an explosive effect during learning due to the power exponent. This pre-processing is, however, adjusted on the training set, the adjustment meaning only the search for the maximum value of each series, which is then used directly on the test set. This means that on the test set, it is possible to have data greater than 1, but as no optimization is used, this is not a problem. Finally, we split our data set into a training set and a test set, with a standard proportion of 80-20. This represents over 21,000 points in the training set and 5,000 in the test set. 

\subsection{Loss Function for Model Training}
Since we have a numerical prediction problem, we have opted to optimize our model using the root mean square error (RMSE) as the loss function, whose formula is simple:
$$
\text{MSE} = \frac{1}{N} \sum_{i=1}^{N} \left(\hat{X}_{t+1}^{(i)} - X_{t+1}^{(i)}\right)^2,
$$
where \(N\) represents the number of samples in the dataset, \(\hat{X}_{t+1}^{(i)}\) denotes the predicted notional values of Bitcoin at time \(t+1\) for the \(i\)-th sample, and \(X_{t+1}^{(i)}\) are the corresponding true values.

The first reason is that this is the most widely used and standard method in machine learning for this type of problem.
The second reason is the metric we want to use to display the results, namely the R-squared \(R^2\). The \(R^2\) is interesting as a metric because it not only gives information on the error but also on the error given the variance of the estimated series, which means it's much easier to tell whether the model is performing well or not. It is also a measure widely used by econometricians and practitioners for this very reason. However, minimizing the MSE is exactly the same as maximizing the \(R^2\), as its formula indicates:
$$
R^2 = 1 - \frac{\sum_{i=1}^{N} (\hat{X}_{t+1}^{(i)} - X_{t+1}^{(i)})^2}{\sum_{i=1}^{N} (X_{t+1}^{(i)} - \bar{X}_{t+1})^2}.
$$
As we can see, the upper terms of the quotient are equivalent to the sum of the squared errors, the other two components being fixed, the optimization of the mean squared error is totally similar to the optimization of $R^2$.

\subsection{Benchmarks}

\subsubsection{Model Architectures}
To assess the merits of our model, we've tested it on various benchmarks. Firstly, we are reusing those of \cite{genet2024tkan}, as this is an important baseline, and here we are also comparing the difference with the variant that includes the TKAN layer versus a variant that does not include it. Finally, we have added a few points of comparison to justify why we removed the top part of a standard temporal fusion-transformation architecture with result tables if the top parts were retained.

\medskip

In order to analyze the performance of our model, here is the model tested:

\begin{enumerate}
	\item TKAT: refers to the paper model implementation, which uses TKAN layers and flattening directly after multi-head attention. We use a number of hidden units of 100 in each layer, with the exception of the embedding layer, which has only one unit per feature. The number of heads used is 4.
	\item TKATN: follows the same implementation, but replaces the TKAN layers with standard LSTM layers to see whether the TKAN layers really improve the model or not.
	\item TKAN, GRU and LSTM: Simple model described in \cite{genet2024tkan} consisting of two 100-unit recurrent layers finalized by a dense linear layer.
    \item TKAT-A, TKATN-A: Variant that adds a Gate, Add and Normalization layer after the multihead with feedforward of the multihead's attention input. This shows why we have removed this part of our implementation.
    \item TKAT-B, TKATN-B: Similar to version A above, with a GRN after the Gate, Add and Normalization, and a final Gate, Add, Normalization with feedforward of the pre-multihead GRN inputs. This architecture is almost entirely similar to the TFT architecture, except that the entire sequence is used after the multihead attention layer, whereas it is only used for the future part in the TFT architecture.
\end{enumerate}
It is to note than we did not use dropout in any above models.

\medskip

\subsubsection{Note on training details}
Such in \cite{genet2024tkan}, Metrics are calculated directly on scaled data and not on unscaled data. There are two reasons for this: firstly, MinMax scaling has no impact on the metric since the minimum is 0 and the data interval is [0,1]; rescaling would simply have involved multiplying all the points, which would not have changed the R-squared.
Nor do we rescale from the median split, as otherwise the series mean would not have been stable over time, leading to a drift in error magnitude on certain parts of the series, rendering the metric insignificant.In terms of optimizing model details, we used the Adam optimizer, one of the most popular choices, used 20\% of our training set as a validation set and included two training recalls. The first is an early learning reminder, which interrupts training after 6 consecutive periods without improvement on the validation set, and restores the weights associated with the best loss obtained on the validation set. The second is a plateau learning rate reduction, which halves the learning rate after 3 consecutive periods showing no improvement on the validation set.

\section{Results}
The results will be illustrated in the latter section, all the methodology, data and codes are available to reproduce the experience. In order to evaluate the model's performance, taking into account the randomness induced by the weight initializer, we repeat the experiment 5 times for each, and display below the mean and standard deviation of the train results obtained from these 5 experiments.

\subsection{Performance metrics TKAT summary vs. benchmarks}

\begin{table}[H]
    \centering
    \caption{\(R^2\) Average: TKAT vs Benchmark}
    \begin{tabular}{cccccc}
    \toprule
    Time & TKAT & TKAT non-KAN & TKAN & GRU & LSTM \\
    \midrule
    1  & 0.30519 & 0.29834 & 0.35084 & 0.36513 & 0.35553 \\
    3  & 0.21801 & 0.22146 & 0.19884 & 0.20067 & 0.06122  \\
    6  & 0.17955 & 0.17584 & 0.14054 & 0.08250 & -0.22583  \\
    9  & 0.16476 & 0.15378 & 0.11747 & 0.08716 & -0.29058  \\
    12 & 0.14908 & 0.15179 & 0.10511 & 0.01786 & -0.47322  \\
    15 & 0.14504 & 0.12658 & 0.08607 & 0.03342 & -0.40443 \\
    \bottomrule
    \end{tabular}
\end{table}

\begin{table}[H]
    \centering
    \caption{\(R^2\) Standard Deviation: TKAT vs Benchmark}
    \begin{tabular}{cccccc}
    \toprule
    Time & TKAT & TKAT non-KAN & TKAN & GRU & LSTM \\
    \midrule
    1  & 0.01886 & 0.01610 & 0.01420 & 0.00833 & 0.01116 \\
    3  & 0.00906 & 0.00485 & 0.00738 & 0.00484 & 0.08020 \\
    6  & 0.00654 & 0.00352 & 0.00723 & 0.02363 & 0.06271 \\
    9  & 0.00896 & 0.00578 & 0.01455 & 0.01483 & 0.05272 \\
    12 & 0.00477 & 0.00507 & 0.00391 & 0.08638 & 0.08574 \\
    15 & 0.01014 & 0.01106 & 0.02465 & 0.02407 & 0.09272 \\
    \bottomrule
    \end{tabular}
\end{table}

Similar to the results obtained with TKAN previously, TKAT shows weaker predictive power on the single-step prediction task compared with a much simpler model, which may be explained by the excessive complexity of the model for this task. However, it appears that as soon as the model is used with only three steps ahead, it is able to offer much better performance than all the others. The results show that the results obtained are more stable during execution, and over the number of time steps with $R^2$ more than 50\% higher than with the simple TKAN model. This makes sense, as the TFT-inspired transformer architecture is much more complete and considered state-of-the-art for these tasks. 

\medskip

Another very interesting result is the comparison between the model using the TKAN layer and the one based on LSTMs. While the architecture of the model makes the LSTM perform much better than on its own, it appears that the TKAN layer is able to improve performance by around 5\% in average. One reading we can therefore make of the results here is that the overall architecture of the model plays an important role in the results obtained, particularly in comparison with the simple model variant, but that the TKAN layers are capable of increasing model performance even in more complex architecture.

\subsection{Performance Metrics Summary of the TKAT versus variant closer to TFT}.

The model, while inspired by the architecture of the temporal fusion transformer, presents some major divergences from it, due to the task we are studying. The first is the absence of static covariates, which are less relevant here, especially as the inclusion of some of them would have completely altered the training tasks themselves and thus made comparison with TKAN performance more difficult.

Secondly, as we changed the whole upper part of the TFT after the multiple head, mainly because our tasks are very different from the ones they used, we felt that this architecture would not match our problem. This difference is mainly due to the fact that the data in the TFT article is data where the known part of the features is much larger, whereas our tasks have many more observed inputs. That said, we still tested variants closer to this article to see if our intuition on this subject was justified or not.
\begin{table}[H]
\centering
\caption{\(R^2\) Average: TKAT variants closer to TFT}
\begin{tabular}{ccccc}
\toprule
Time & TKAT-A & TKATN-A & TKAT-B & TKATN-B \\
\midrule
1  & 0.29237 & 0.29502 & 0.28861 & 0.31119 \\
3  & 0.19352 & 0.11464 & 0.16413 & 0.12965 \\
6  & 0.11667 & 0.11588 & 0.14149 & 0.12075  \\
9  & 0.07502 & 0.10003 & 0.09655 & 0.05756 \\
12 & 0.05745 & 0.06382 & 0.07805 & 0.05968 \\
15 & 0.04640 & 0.03845 & 0.05999 & 0.04139 \\
\bottomrule
\end{tabular}
\end{table}

\begin{table}[H]
\centering
\caption{\(R^2\) Standard Deviation: TKAT variants closer to TFT}
\begin{tabular}{ccccc}
\toprule
Time & TKAT-A & TKATN-A & TKAT-B & TKATN-B \\
\midrule
1  & 0.04861 & 0.03480 & 0.08500 & 0.00833 \\
3  & 0.02206 & 0.04145 & 0.02807 & 0.02665 \\
6  & 0.02927 & 0.02070 & 0.01188 & 0.01567 \\
9  & 0.01558 & 0.02216 & 0.01776 & 0.01475\\
12 & 0.01437 & 0.01354 & 0.01431 & 0.02162 \\
15 & 0.01191 & 0.00834 & 0.01612 & 0.00937 \\
\bottomrule
\end{tabular}
\end{table}
In view of these results, it's more than obvious that using the full architecture would lead to poorer results for our specific tasks. The architecture we propose is therefore clearly different from that of TFT, but is probably not adapted to the same tasks by construction. What we want to highlight is that while the TFT inspiration gives excellent results for improving the performance of simple models, it needs to be adapted to specific learning tasks, as here the results show not only a lower mean $R^2$ but also much more variance in the results.

\subsection{Number of parameters per model}
\medskip
A metric often looked at in Deep-Learning architectures is the number of parameters used by them. Here is the number of trainable parameters for the above models, calculated by tensorflow 2.16.

\begin{table}[H]
    \centering
    \caption{Number of weights per model}
    \begin{tabular}{cccccc}
    \toprule
    Time & TKAT & TKAT non-KAN & TKAN & GRU & LSTM \\
    \midrule
    1  & 1,022,275 & 1,061,461 & 99,426 & 97,001 & 128,501 \\
    3  & 1,027,077 & 1,066,263 & 99,628 & 97,203 & 128,703 \\
    6  & 1,043,280 & 1,082,466 & 99,931 & 97,506 & 129,006 \\
    9  & 1,070,283 & 1,109,469 & 100,234 & 97,809 & 129,309 \\
    12 & 1,108,086 & 1,147,272 & 100,537 & 98,112 & 129,612 \\
    15 & 1,156,689 & 1,195,875 & 100,840 & 98,415 & 129,915 \\
    \bottomrule
    \end{tabular}
\end{table}

Clearly, while TKAN's layers are not very different from standard RNNs such as GRU and LSTM, TKAT displays almost ten times as many parameters due to its more complex architecture. It's worth noting, however, that almost 65\% of these trainable parameters belong to the variable selection part of the network, whereas the TKAN/LSTM layers only account for 10\% of the total trainable weights. This metric, while interesting in some respects, should be taken with caution, as two models with the same number of parameters can have very different learning times, and the KAN part is, for the same number of parameters, more difficult to train than a standard MLP. 

\subsection{Results over Steps on 30 steps forecasting task}
\medskip
Finally, we also wanted to display the root-mean-square error for each forward step as a function of a model, all trained to predict the next 30 periods, because although the $R^2$ is readable, it gives no information on errors over the steps. All the models are those described above, with one extra, the MLP. This flattens the inputs received and simply applies two dense layers of 100 units with relu activation, before a final dense linear layer of 30 units to match the number of steps to be predicted. 

The models therefore have the following number of parameters:
\begin{table}[H]
    \centering
    \caption{Trainable Parameters Count}
    \begin{tabular}{ccc}
    \toprule
    Model & Number of Trainable Parameters & \(R^2\)\\
    \midrule
    TKAT  & 1,561,704 & 0.127743 \\
    TKAN  & 102,355 & 0.068004 \\
    MLP  & 298,230 & -0.036020 \\
    GRU  & 99,930 & 0.055263 \\
    LSTM & 131,430 & -0.008245 \\
    \bottomrule
    \end{tabular}
\end{table}

The results were obtained after a single run for this graph.

\begin{figure}[H]
    \centering
    \includegraphics[width=0.7\linewidth]{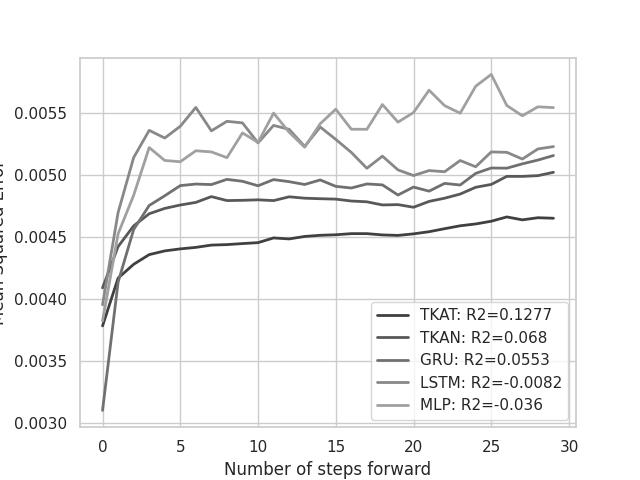}
    \caption{Model and errors based on number of steps forward} 
    \label{fig:result_graph}
\end{figure}

The graph clearly shows the clear outperformance of GRU in the first step, in line with previous results, but also the fact that when there are more steps, TKAT and TKAN improve the quality of the forecast. 

\section{Conclusion}
In this article, we have proposed an adaptation of two cutting-edge concepts in the field of deep learning, namely the Transformer architecture and the recent Kolmogorov-Arnold network through its version. The results obtained show a real improvement over standard methods and demonstrate that TKAN is something to consider even in Transformer architecture. With our specific learning task, different from that used for the Temporal Fusion Transformer, we show that the architecture needs to be modified task by task to achieve better performance, and that ours clearly outperforms the standard method, even though it lacks the flexibility to use static covariates. It's also worth noting that this research has shown that the main architecture of the model plays a more important role than the single layer in performance.  These results are very promising and confirm those of the TKAN paper.

\bibliographystyle{IEEEtran}
\bibliography{bib}

\begin{thebibliography}{10}
\providecommand{\url}[1]{#1}
\csname url@samestyle\endcsname
\providecommand{\newblock}{\relax}
\providecommand{\bibinfo}[2]{#2}
\providecommand{\BIBentrySTDinterwordspacing}{\spaceskip=0pt\relax}
\providecommand{\BIBentryALTinterwordstretchfactor}{4}
\providecommand{\BIBentryALTinterwordspacing}{\spaceskip=\fontdimen2\font plus
\BIBentryALTinterwordstretchfactor\fontdimen3\font minus \fontdimen4\font\relax}
\providecommand{\BIBforeignlanguage}[2]{{%
\expandafter\ifx\csname l@#1\endcsname\relax
\typeout{** WARNING: IEEEtran.bst: No hyphenation pattern has been}%
\typeout{** loaded for the language `#1'. Using the pattern for}%
\typeout{** the default language instead.}%
\else
\language=\csname l@#1\endcsname
\fi
#2}}
\providecommand{\BIBdecl}{\relax}
\BIBdecl

\bibitem{taieb2010multiple}
S.~B. Taieb, A.~Sorjamaa, and G.~Bontempi, ``Multiple-output modeling for multi-step-ahead time series forecasting,'' \emph{Neurocomputing}, vol.~73, no. 10-12, pp. 1950--1957, 2010.

\bibitem{tiao1989model}
G.~C. Tiao and R.~S. Tsay, ``Model specification in multivariate time series,'' \emph{Journal of the Royal Statistical Society: Series B (Methodological)}, vol.~51, no.~2, pp. 157--195, 1989.

\bibitem{wei2018multivariate}
W.~W. Wei, \emph{Multivariate time series analysis and applications}.\hskip 1em plus 0.5em minus 0.4em\relax John Wiley \& Sons, 2018.

\bibitem{marcellino2006comparison}
M.~Marcellino, J.~H. Stock, and M.~W. Watson, ``A comparison of direct and iterated multistep ar methods for forecasting macroeconomic time series,'' \emph{Journal of econometrics}, vol. 135, no. 1-2, pp. 499--526, 2006.

\bibitem{zivot2006vector}
E.~Zivot and J.~Wang, ``Vector autoregressive models for multivariate time series,'' \emph{Modeling financial time series with S-PLUS{\textregistered}}, pp. 385--429, 2006.

\bibitem{lutkepohl2013vector}
H.~L{\"u}tkepohl, ``Vector autoregressive models,'' in \emph{Handbook of research methods and applications in empirical macroeconomics}.\hskip 1em plus 0.5em minus 0.4em\relax Edward Elgar Publishing, 2013, pp. 139--164.

\bibitem{hamilton1994state}
J.~D. Hamilton, ``State-space models,'' \emph{Handbook of econometrics}, vol.~4, pp. 3039--3080, 1994.

\bibitem{kim1994dynamic}
C.-J. Kim, ``Dynamic linear models with markov-switching,'' \emph{Journal of econometrics}, vol.~60, no. 1-2, pp. 1--22, 1994.

\bibitem{tang1991time}
Z.~Tang, C.~De~Almeida, and P.~A. Fishwick, ``Time series forecasting using neural networks vs. box-jenkins methodology,'' \emph{Simulation}, vol.~57, no.~5, pp. 303--310, 1991.

\bibitem{binkowski2018autoregressive}
M.~Binkowski, G.~Marti, and P.~Donnat, ``Autoregressive convolutional neural networks for asynchronous time series,'' in \emph{International Conference on Machine Learning}.\hskip 1em plus 0.5em minus 0.4em\relax PMLR, 2018, pp. 580--589.

\bibitem{ma2019novel}
K.~Ma and H.~Leung, ``A novel lstm approach for asynchronous multivariate time series prediction,'' in \emph{2019 International Joint Conference on Neural Networks (IJCNN)}.\hskip 1em plus 0.5em minus 0.4em\relax IEEE, 2019, pp. 1--7.

\bibitem{hochreiter1997long}
S.~Hochreiter and J.~Schmidhuber, ``Long short-term memory,'' \emph{Neural computation}, vol.~9, no.~8, pp. 1735--1780, 1997.

\bibitem{vaswani2017attention}
A.~Vaswani, N.~Shazeer, N.~Parmar, J.~Uszkoreit, L.~Jones, A.~N. Gomez, {\L}.~Kaiser, and I.~Polosukhin, ``Attention is all you need,'' \emph{Advances in neural information processing systems}, vol.~30, 2017.

\bibitem{shih2019temporal}
S.-Y. Shih, F.-K. Sun, and H.-y. Lee, ``Temporal pattern attention for multivariate time series forecasting,'' \emph{Machine Learning}, vol. 108, pp. 1421--1441, 2019.

\bibitem{wan2019multivariate}
R.~Wan, S.~Mei, J.~Wang, M.~Liu, and F.~Yang, ``Multivariate temporal convolutional network: A deep neural networks approach for multivariate time series forecasting,'' \emph{Electronics}, vol.~8, no.~8, p. 876, 2019.

\bibitem{liu2024kan}
Z.~Liu, Y.~Wang, S.~Vaidya, F.~Ruehle, J.~Halverson, M.~Solja{\v{c}}i{\'c}, T.~Y. Hou, and M.~Tegmark, ``Kan: Kolmogorov-arnold networks,'' \emph{arXiv preprint arXiv:2404.19756}, 2024.

\bibitem{vaca2024kolmogorov}
C.~J. Vaca-Rubio, L.~Blanco, R.~Pereira, and M.~Caus, ``Kolmogorov-arnold networks (kans) for time series analysis,'' \emph{arXiv preprint arXiv:2405.08790}, 2024.

\bibitem{genet2024tkan}
R.~Genet and H.~Inzirillo, ``Tkan: Temporal kolmogorov-arnold networks,'' \emph{arXiv preprint arXiv:2405.07344}, 2024.

\bibitem{bozorgasl2024wav}
Z.~Bozorgasl and H.~Chen, ``Wav-kan: Wavelet kolmogorov-arnold networks,'' \emph{arXiv preprint arXiv:2405.12832}, 2024.

\bibitem{lim2021temporal}
B.~Lim, S.~{\"O}. Ar{\i}k, N.~Loeff, and T.~Pfister, ``Temporal fusion transformers for interpretable multi-horizon time series forecasting,'' \emph{International Journal of Forecasting}, vol.~37, no.~4, pp. 1748--1764, 2021.

\bibitem{mehta2020delight}
S.~Mehta, M.~Ghazvininejad, S.~Iyer, L.~Zettlemoyer, and H.~Hajishirzi, ``Delight: Deep and light-weight transformer,'' \emph{arXiv preprint arXiv:2008.00623}, 2020.

\bibitem{raffel2020exploring}
C.~Raffel, N.~Shazeer, A.~Roberts, K.~Lee, S.~Narang, M.~Matena, Y.~Zhou, W.~Li, and P.~J. Liu, ``Exploring the limits of transfer learning with a unified text-to-text transformer,'' \emph{Journal of machine learning research}, vol.~21, no. 140, pp. 1--67, 2020.

\bibitem{so2019evolved}
D.~So, Q.~Le, and C.~Liang, ``The evolved transformer,'' in \emph{International conference on machine learning}.\hskip 1em plus 0.5em minus 0.4em\relax PMLR, 2019, pp. 5877--5886.

\bibitem{fan2021mask}
Z.~Fan, Y.~Gong, D.~Liu, Z.~Wei, S.~Wang, J.~Jiao, N.~Duan, R.~Zhang, and X.~Huang, ``Mask attention networks: Rethinking and strengthen transformer,'' \emph{arXiv preprint arXiv:2103.13597}, 2021.

\bibitem{shoeybi2019megatron}
M.~Shoeybi, M.~Patwary, R.~Puri, P.~LeGresley, J.~Casper, and B.~Catanzaro, ``Megatron-lm: Training multi-billion parameter language models using model parallelism,'' \emph{arXiv preprint arXiv:1909.08053}, 2019.

\bibitem{dai1901attentive}
Z.~Dai, Z.~Yang, Y.~Yang, J.~Carbonell, Q.~Le, and R.~T.-X. Salakhutdinov, ``Attentive language models beyond a fixed-length context. published online june 2, 2019. doi: 10.48550,'' \emph{arXiv}, 1901.

\bibitem{roy2021efficient}
A.~Roy, M.~Saffar, A.~Vaswani, and D.~Grangier, ``Efficient content-based sparse attention with routing transformers,'' \emph{Transactions of the Association for Computational Linguistics}, vol.~9, pp. 53--68, 2021.

\bibitem{rae2019compressive}
J.~W. Rae, A.~Potapenko, S.~M. Jayakumar, and T.~P. Lillicrap, ``Compressive transformers for long-range sequence modelling,'' \emph{arXiv preprint arXiv:1911.05507}, 2019.

\bibitem{yan2019tener}
H.~Yan, B.~Deng, X.~Li, and X.~Qiu, ``Tener: adapting transformer encoder for named entity recognition,'' \emph{arXiv preprint arXiv:1911.04474}, 2019.

\bibitem{arkhipov2019tuning}
M.~Arkhipov, M.~Trofimova, Y.~Kuratov, and A.~Sorokin, ``Tuning multilingual transformers for language-specific named entity recognition,'' in \emph{Proceedings of the 7th Workshop on Balto-Slavic Natural Language Processing}, 2019, pp. 89--93.

\bibitem{yu2020improving}
J.~Yu, J.~Jiang, L.~Yang, and R.~Xia, ``Improving multimodal named entity recognition via entity span detection with unified multimodal transformer.''\hskip 1em plus 0.5em minus 0.4em\relax Association for Computational Linguistics, 2020.

\bibitem{wu2020deep}
N.~Wu, B.~Green, X.~Ben, and S.~O'Banion, ``Deep transformer models for time series forecasting: The influenza prevalence case,'' \emph{arXiv preprint arXiv:2001.08317}, 2020.

\bibitem{zhou2021informer}
H.~Zhou, S.~Zhang, J.~Peng, S.~Zhang, J.~Li, H.~Xiong, and W.~Zhang, ``Informer: Beyond efficient transformer for long sequence time-series forecasting,'' in \emph{Proceedings of the AAAI conference on artificial intelligence}, vol.~35, no.~12, 2021, pp. 11\,106--11\,115.

\bibitem{li2019enhancing}
S.~Li, X.~Jin, Y.~Xuan, X.~Zhou, W.~Chen, Y.-X. Wang, and X.~Yan, ``Enhancing the locality and breaking the memory bottleneck of transformer on time series forecasting,'' \emph{Advances in neural information processing systems}, vol.~32, 2019.

\bibitem{wu2020adversarial}
S.~Wu, X.~Xiao, Q.~Ding, P.~Zhao, Y.~Wei, and J.~Huang, ``Adversarial sparse transformer for time series forecasting,'' \emph{Advances in neural information processing systems}, vol.~33, pp. 17\,105--17\,115, 2020.

\bibitem{kolmogorov1961representation}
A.~N. Kolmogorov, \emph{On the representation of continuous functions of several variables by superpositions of continuous functions of a smaller number of variables}.\hskip 1em plus 0.5em minus 0.4em\relax American Mathematical Society, 1961.

\bibitem{staudemeyer2019understanding}
R.~C. Staudemeyer and E.~R. Morris, ``Understanding lstm--a tutorial into long short-term memory recurrent neural networks,'' \emph{arXiv preprint arXiv:1909.09586}, 2019.

\bibitem{clevert2015fast}
D.-A. Clevert, T.~Unterthiner, and S.~Hochreiter, ``Fast and accurate deep network learning by exponential linear units (elus),'' \emph{arXiv preprint arXiv:1511.07289}, 2015.

\bibitem{ba2016layer}
J.~L. Ba, J.~R. Kiros, and G.~E. Hinton, ``Layer normalization,'' 2016.

\bibitem{dauphin2017language}
Y.~N. Dauphin, A.~Fan, M.~Auli, and D.~Grangier, ``Language modeling with gated convolutional networks,'' in \emph{International conference on machine learning}.\hskip 1em plus 0.5em minus 0.4em\relax PMLR, 2017, pp. 933--941.

\end{thebibliography}
\end{document}